\documentclass[runningheads]{llncs}
\usepackage{graphicx}
\usepackage{comment}
\usepackage{amsmath,amssymb} 
\usepackage{color}
\usepackage{soul}
\usepackage[compatibility=false]{caption}
\usepackage{subcaption}
\usepackage{multirow}
\usepackage{dcolumn}
\usepackage{booktabs}
\usepackage[pagebackref=true,breaklinks=true,letterpaper=true,colorlinks,bookmarks=false]{hyperref}
\usepackage{cleveref}

\begin{document}
\pagestyle{headings}
\mainmatter
\def\ECCVSubNumber{5635}  

\title{All at Once: Temporally Adaptive Multi-Frame Interpolation with Advanced Motion Modeling} 

\titlerunning{All at Once}
%
\author{Zhixiang Chi\inst{1},
Rasoul Mohammadi Nasiri\inst{1},
Zheng Liu\inst{1},
Juwei Lu\inst{1},
Jin Tang\inst{1}
Konstantinos N Plataniotis\inst{2} \orcidID{0000-0003-3647-5473}}
\authorrunning{Z. Chi et al.}
%

\institute{$^{1}$Noah’s Ark Lab, Huawei Technologies \quad $^{2}$University of Toronto, Canada\\
\email{\{zhixiang.chi, rasoul.nasiri, zheng.liu1, tangjin, juwei.lu\}@huawei.com, kostas@ece.utoronto.ca}
}


\maketitle

\begin{abstract}

Recent advances in high refresh rate displays as well as the increased interest in high rate of slow motion and frame up-conversion fuel the demand for efficient and cost-effective multi-frame video interpolation solutions.  To that regard, inserting multiple frames between consecutive video frames are of paramount importance for the consumer electronics industry. State-of-the-art methods are iterative solutions interpolating one frame at the time. They introduce temporal inconsistencies and clearly noticeable visual artifacts. 

Departing from the state-of-the-art, this work introduces a true multi-frame interpolator. It utilizes a pyramidal style network in the temporal domain to complete the multi-frame interpolation task in one-shot. A novel flow estimation procedure using a relaxed loss function, and an advanced, cubic-based, motion model is also used to further boost interpolation accuracy when complex motion segments are encountered. Results on the Adobe240 dataset show that the proposed method generates visually pleasing, temporally consistent frames, outperforms the current best off-the-shelf method by 1.57db in PSNR with 8 times smaller model and 7.7 times faster. The proposed method can be easily extended to interpolate a large number of new frames while remaining efficient because of the one-shot mechanism. \href{https://chi-chi-zx.github.io/all-at-once/}{https://chi-chi-zx.github.io/all-at-once/}

\end{abstract}

\section{Introduction}

Video frame interpolation targets generating new frames for the moments in which no frame is recorded. It is mostly used in slow motion generation~\cite{xu2019quadratic}, adaptive streaming~\cite{wu2015modeling}, and frame rate up-conversion~\cite{castagno1996method}. The fast innovation in high refresh rate displays and great interests in a higher rate of slow motion and frame up-conversion bring the needs to multi-frame interpolation.

Recent efforts focus on the main challenges of interpolation, including occlusion and large motions, but they have not explored the temporal consistency as a key factor in video quality, especially for multi-frame interpolation. Almost all the existing methods interpolate one frame in each execution, and generating multiple frames can be addressed by either iteratively generating a middle frame~\cite{niklaus2017video,liu2017video,xue2019video} or independently creating each intermediate frame for corresponding time stamp~\cite{jiang2018super,DAIN,bao2018memc,niklaus2018context,liu2019deep}. The former approach might cause error propagation by treating the generated middle frame as input. As well, the later one may suffer from temporal inconsistency due to the independent process for each frame and causes temporal jittering at playback. Those artifacts are further enlarged when more frames are interpolated. An important point that has been missed in existing methods is the variable level of difficulties in generating intermediate frames. In fact, the frames closer to the two initial frames are easier to generate, and those with larger temporal distance are more difficult. Consequently, the current methods are not optimized in terms of model size and running time for multi-frame interpolation, which makes them inapplicable for real-life applications. 

On the other hand, most of the state-of-the-art interpolation methods commonly synthesize the intermediate frames by simply assuming linear transition in motion between the pair of input frames. However, real-world motions reflected in video frames follow a variety of complex non-linear trends~\cite{xu2019quadratic}. While a quadratic motion prediction model is proposed in~\cite{xu2019quadratic} to overcome this limitation, it is still inadequate to model real-world scenarios especially for non-rigid bodies, by assuming constant acceleration. As forces applied to move objects in the real world are not necessarily constant, it results in variation in acceleration.

To this end, we propose a temporal pyramidal processing structure that efficiently integrates the multi-frame generation into one single network. Based on the expected level of difficulties, we adaptively process the easier cases~(frames) with shallow parts to guide the generation of harder frames that are processed by deeper structures. Through joint optimization of all the intermediate frames, higher quality and temporal consistency can be ensured. In addition, we exploit the advantage of multiple input frames as in~\cite{xu2019quadratic,lee2013frame} to propose an advanced higher-order motion prediction modeling, which explores the variation in acceleration. Furthermore, inspired by~\cite{xue2019video}, we develop a technique to boost the quality of motion prediction as well as the final interpolation results by introducing a relaxed loss function to the optical flow~(O.F.) estimation module. In particular, it gives the flexibility to map the pixels to the neighbor of their ground truth locations at the reference frame while a better motion prediction for the intermediate frames can be achieved. Comparing to the current state-of-the-art method~\cite{xu2019quadratic}, we outperform it in interpolation quality measured by PSNR by 1.57dB on the Adobe240 dataset and achieved 8 times smaller in model size and 7.7 times faster in generating 7 frames.

We summarize our contributions as 1) We propose a temporal pyramidal structure to integrate the multi-frame interpolation task into one single network to
generate temporally consistent and high-quality frames; 2) We propose a higher-order motion modeling to exploit variations in acceleration involved in real-world motion; 3) We develop a relaxed loss function to the flow estimation task to boost the interpolation quality; 4) We optimize the network size and speed so that it is applicable for the real world applications especially for mobile devices.


\section{Related work}

Recent efforts on frame interpolation have focused on dealing with the main sources of degradation in interpolation quality, such as large motion and occlusion. Different ideas have been proposed such as estimating occlusion maps~\cite{jiang2018super,yuan2019zoom}, learning adaptive kernel for each pixel~\cite{niklaus2017video,niklaus2017video_1}, exploring depth information~\cite{DAIN} or extracting deep contextual features~\cite{niklaus2018context,bao2018memc}. As most of these methods interpolate frames one at a time, inserting multiple frames is achieved by iteratively executing the models. In fact, as a fundamental issue, the step-wise implementation of multi-frame interpolation does not consider the time continuity and may cause temporally inconsistency. In contrast, generating multiple frames in one integrated network will implicitly enforce the network to generate temporally consistent sequences. 
The effectiveness of the integrated approach has been verified by Super~SloMo~\cite{jiang2018super}; however, their method is not purposely designed for the task of multi-frame interpolation. Specifically, what has been missed in~\cite{jiang2018super} is to utilize the error cue from temporal distance between a middle frame and the input frames and optimize the whole model accordingly. Therefore, the proposed adaptive processing based on this difficulty pattern can result in a more optimized solution, which is not considered in the state-of-the-art methods~\cite{jiang2018super,DAIN,bao2018memc,xu2019quadratic,niklaus2017video}.

Given the estimated O.F. among the input frames, one important step in frame interpolation is modeling the traversal of pixels in between the two frames. The most common approach is to consider a linear transition and scaling of the O.F.~\cite{yuan2019zoom,niklaus2018context,jiang2018super,liu2017video,bao2018memc,DAIN}. Recent work in~\cite{xu2019quadratic,bao2018high} applied an acceleration-aware method by also contributing the neighborhood frames of the initial pair. However, in real life, the force applied to the moving object is not constant; thus, the motion is not following the linear or quadratic pattern. In this paper, we propose a simple but powerful higher-order model to handle more complex motions happen in the real world and specially non-rigid bodies. On the other hand, \cite{jiang2018super} imposes accurate estimation the O.F. by the warping loss. However, \cite{xue2019video} reveals that accurate O.F. is not tailored for task-oriented problems. Motivated by that, we apply a flexible O.F. estimation between initial frames, which gives higher flexibility to model complex motions.

\section{Proposed method}

\subsection{Algorithm overview}
\begin{figure*}[t]
    \centering
    \includegraphics[width =\linewidth]{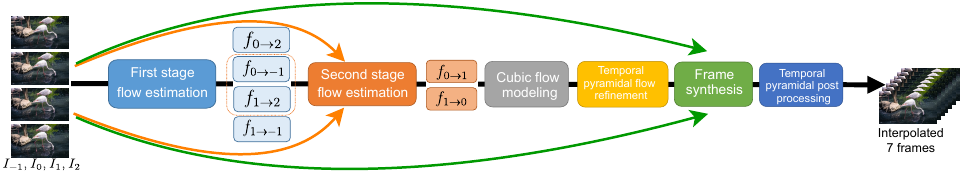}
    \caption{An overview of the proposed multi-frame interpolation method.}
    \label{fig:overall}
\end{figure*}

An overview of the proposed method is shown in \figurename~\ref{fig:overall} where we use four input frames ($I_{-1}, I_0, I_1$ and $I_2$) to generate 7 frames ($I_{t_i}, t_i=\frac{i}{8},i\in[1,2,\cdots,7]$) between $I_0$ and $I_1$. We first use two-step O.F. estimation module to calculate O.F.s ($f_{0 \rightarrow 1},f_{1 \rightarrow 0},f_{1 \rightarrow -1},f_{0 \rightarrow 2}$) and then use these flows and cubic modeling to predict the flow between input frames and the new frames. Our proposed temporal pyramidal network then refines the predicted O.F. and generates an initial estimation of middle frames. Finally, the post processing network further improves the quality of interpolated frames ($I_{t_i}$) with the similar temporal pyramid.

\subsection{Cubic flow prediction}
In this work, we integrate the cubic motion modeling to specifically handle the acceleration variation in motions. Considering the motion starting from $I_0$ to a middle time stamp $t_i$ as $f_{0 \rightarrow t_i}$, we model object motion by the cubic model as:
\begin{equation}
\label{eq:cubic}
    f_{0 \rightarrow t_i} = v_0\times t_i + \frac{a_0}{2}\times t^2_i + \frac{\Delta a_0 }{6}\times t^3_i,
\end{equation}
where $v_0$, $a_0$, and $\Delta a_0$ are the velocity, acceleration, and acceleration change rate estimated at $I_0$, respectively. The acceleration terms can be computed as:
\begin{equation}
    \Delta a_0 = a_1-a_0, a_0=f_{0 \rightarrow 1}+f_{0 \rightarrow -1}, a_1=f_{1 \rightarrow 2}+f_{1 \rightarrow 0}.
\end{equation}
where $a_0$ and $a_1$ are calculated for pixels at $I_0$ and $I_1$ respectively. However, the $\Delta a_0$ should be calculated for the pixels correspond to the same real-world point rather than pixels with the same coordinate in the two frames. Therefore, we reformulate $a_1$ to calculated $\Delta a_0$ based on referencing pixel's locations at $I_0$ as:
\begin{equation}
     a_1=f_{0 \rightarrow 2}-2 \times f_{0 \rightarrow 1}.
\end{equation}
To calculate $v_0$ in \eqref{eq:cubic}, the calculation
in~\cite{xu2019quadratic} does not hold when the acceleration is variable, instead, we apply~\eqref{eq:cubic} for $t_i=1$ to solve for $v_0$ using only the information computed above
\begin{equation}
     v_0=f_{0 \rightarrow 1} - \frac{a_0}{2} - \frac{a_1 - a_0}{6}.
\end{equation}
Finally, $f_{0 \rightarrow t_i}$ for any $t_i\in [0,1]$ can be expressed based on only O.F. between input frames by 
\begin{equation}
\label{eq:middle_of}
\small
f_{0 \rightarrow t_i}=f_{0 \rightarrow 1}\times t_i+ \frac{a_0}{2}\times (t_i^2-t_i) + \frac{a_1 - a_0}{6} \times(t_i^3 - t_i).
\end{equation}
$f_{1 \rightarrow t_i}$ can be computed using the same manner. The detailed derivation and proof of all the above equations will be provided in the supplementary document. 

In \figurename~\ref{fig:cubic illustration}, we simulate three different 1-D motions, including constant velocity, constant acceleration, and variable acceleration, as distinguished in three path lines. For each motion, the object position at four time stamps of [$t_0$,$t_1$,$t_2$,$t_3$] are given as shown by gray circles; we apply three predictive models: linear, quadratic\cite{xu2019quadratic} and our cubic model to estimate the location of the object for time stamp $t_{1.5}$ blindly (without having the parameters of simulated motions). The prediction results show that our cubic model is more robust to simulate different order of motions.

\begin{figure}[t!]
    \centering
    \includegraphics[width =\linewidth]{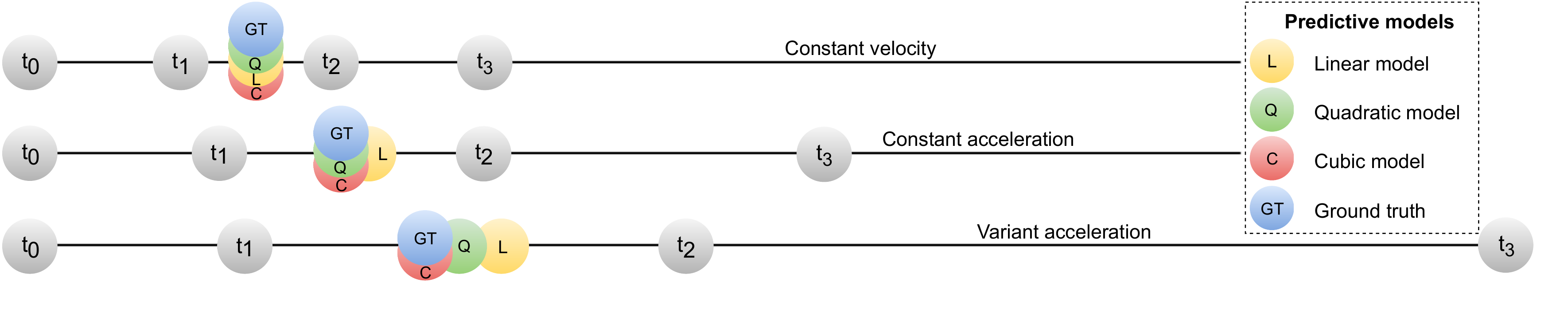}
    \caption{A toy example to illustrate the performance of three models (Linear, Quadratic, and Cubic) in predicting three motion patterns (constant velocity, constant acceleration, and variant acceleration).}
    \label{fig:cubic illustration}
\end{figure}

\subsection{Motion estimation}\label{motion estimation}

\noindent\textbf{Flow estimation module.} To estimate the O.F. among the input frames, the existing frame interpolation methods commonly adopt the off-the-shelf networks~\cite{xu2019quadratic,niklaus2018context,bao2018memc,DAIN,sun2018pwc,dosovitskiy2015flownet,ilg2017flownet}. However, the existing flow networks are not efficiently designed for multi-frame input, and some are limited to one-directional flow estimation. To this end, following the three-scale coarse-to-fine architecture in SPyNet~\cite{ranjan2017optical}, we design a customized two-stage flow estimation to involve the neighbor frames in better estimating O.F. between $I_0$ and $I_1$. Both stages are following similar three-scale architecture, and they optimally share the weights of two coarser levels. The first stage network is designed to compute O.F. between two consecutive frames. We use that to estimate $f_{0 \rightarrow -1}$ and $f_{1 \rightarrow 2}$. In the finest level of second-stage network, we use $I_0$ and $I_1$ concatenated with $-f_{0 \rightarrow -1}$ and $-f_{1 \rightarrow 2}$ as initial estimations to compute $f_{0 \rightarrow 1}$ and $f_{1 \rightarrow 0}$. Alongside, we are calculating the estimation of $f_{0 \rightarrow 2}$ and $f_{1 \rightarrow -1}$ in the first stage, which are used in our cubic motion modeling in later steps.


\noindent\textbf{Motion estimation constraint relaxation.} Common O.F. estimation methods try to map the pixel from the first frame to the exact corresponding location in the second frame. However, TOFlow~\cite{xue2019video} reveals that the accurate O.F. as a part of a higher conceptual level task like frame interpolation does not lead to the optimal solution of that task, especially for occlusion. Similarly, we observed that a strong constraint on O.F. estimation among input frames might degrade the motion prediction for the middle frames, especially for complex motion. In contrast, accepting some flexibility in flow estimation will provide a closer estimation to ground truth motion between frames. The advantage of this flexibility will be illustrated in the following examples.


\begin{figure}[t]
    \centering
    \begin{subfigure}{0.49\linewidth}
    \includegraphics[width =\linewidth]{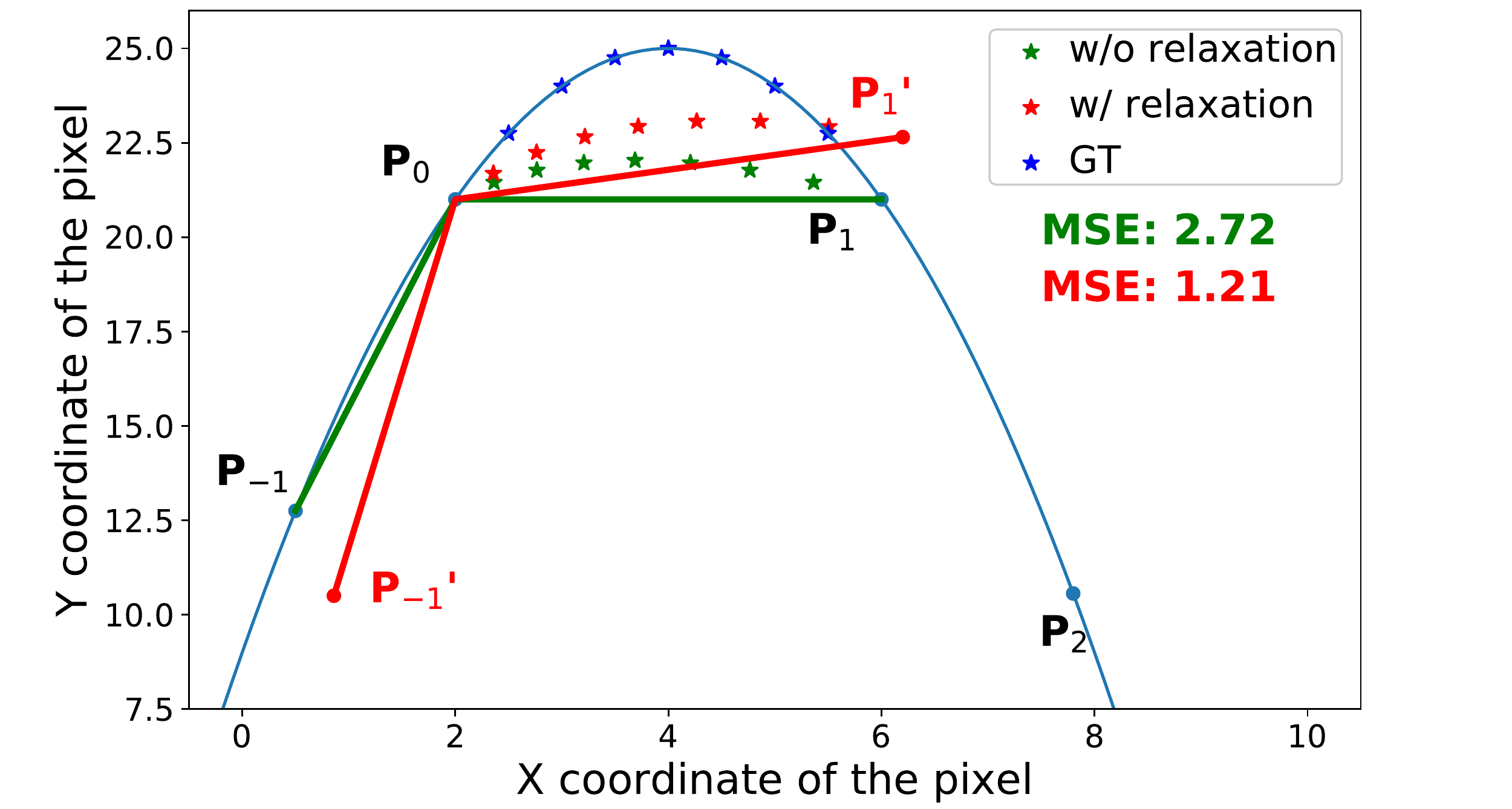}
    \caption{Quadratic prediction.}
    \label{fig: relax_qua}
    \end{subfigure}
    \begin{subfigure}{0.49\linewidth}
    \includegraphics[width =\linewidth]{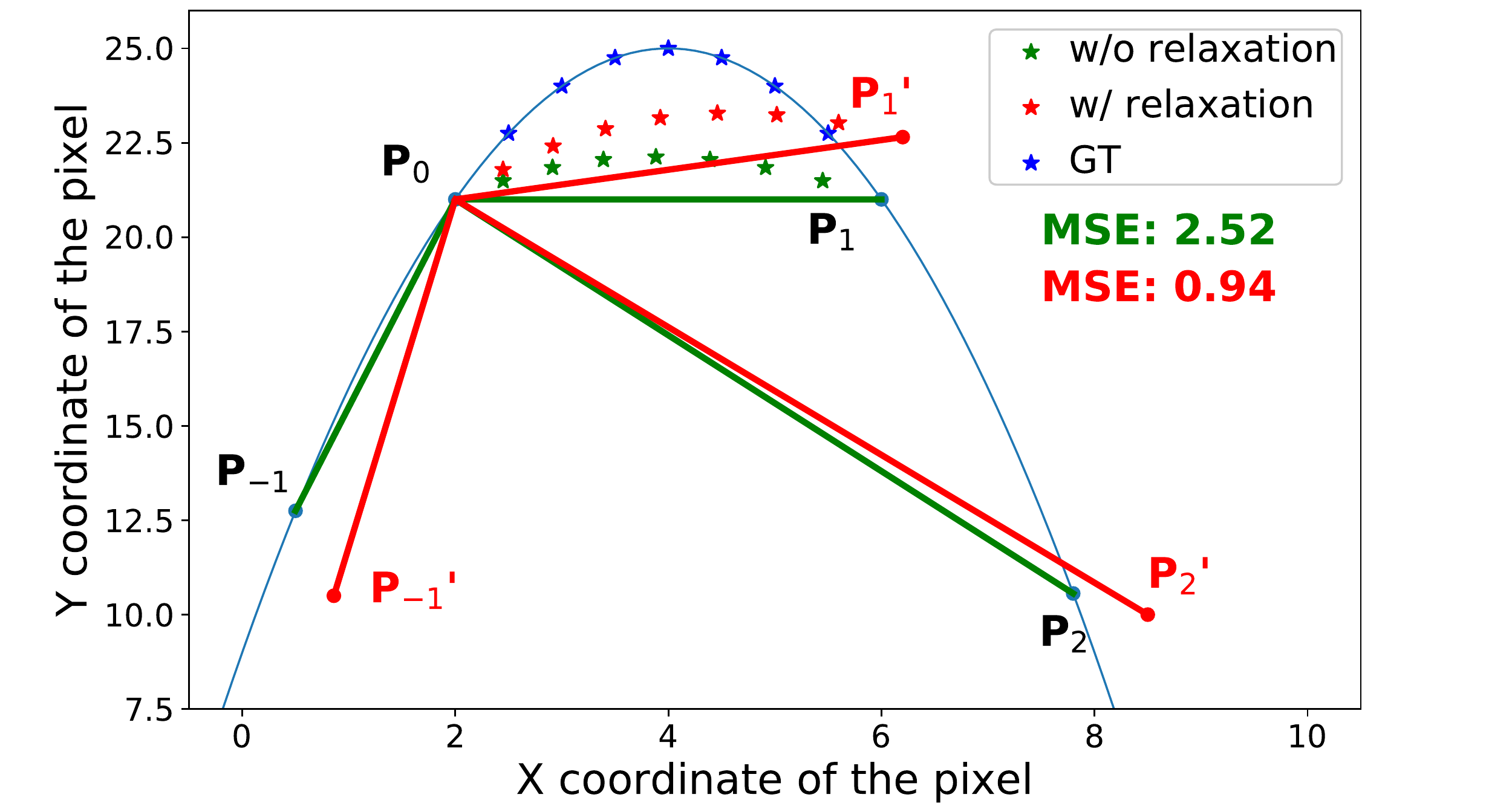}
    \caption{Cubic prediction.}
    \label{fig: relax_cubic}
    \end{subfigure}
    \caption{An example of an object motion path (blue curve) and the motion prediction (with and without relaxation) by Quadratic (a) and Cubic (b) model. 
    }
    \label{fig:relaxation}
\end{figure}

Consider the two toy examples, as shown in \figurename~\ref{fig:relaxation}, where a pixel is moving on the blue curve in consecutive frames and (x,y) is the pixel coordinate in frame space. The pixel position is given in four consecutive frames as $P_{-1}, P_0, P_1$ and $P_2$ and the aim is to find locations for seven moments between $P_0$ and $P_1$ indicated by blue stars. We consider $P_0$ as a reference point in motion prediction. The green lines represent ground truth O.F. between $P_0$ and other points. We predict middle points (green stars) by quadratic~\cite{xu2019quadratic} and cubic models in~\eqref{eq:middle_of} as shown in \figurename~\ref{fig:relaxation}. 
The predicted locations are far from the ground truths (blue stars). However, instead of estimating the exact O.F., giving it a flexibility of mapping $P_0$ to the neighbor of other points denoted as $P_{-1}^{'}$, $P_1^{'}$, $P_2^{'}$, a better prediction of the seven middle locations can be achieved as shown by the red stars. It also reduces the mean squared error~(MSE) significantly. The idea is an analogy to introduce certain errors to the flow estimation process.

To apply the idea of relaxation, we employ the same unsupervised learning in O.F. estimation as~\cite{jiang2018super}, but with a relaxed warping loss. For example, the loss for estimating $f_{0 \rightarrow 1}$ is defined as:
\begin{equation}
        \mathcal{L}^{f_{0 \rightarrow 1}}_{w_{relax}} = \sum_{i=0}^{h-1}\sum_{j=0}^{z-1}\min_{m,n}\left \| I_0^{w \rightarrow 1} (i, j)-I_1(i+m, j+n) \right \|_1,  \mbox{for } m,n\in [-d, +d],
        \label{eq:relaxation}
\end{equation}
where $I_0^{w \rightarrow 1}$ denotes $I_0$ warped by $f_{0 \rightarrow 1}$ to the reference point $I_1$, $d$ determines the range of neighborhood and $h$, $z$ are the image height and width. We use $\mathcal{L}_{w_{relax}}$ for both stages of O.F. estimation. We evaluate the trade-off between the performance of flow estimation and the final results in Section~\ref{ablation section}.

\subsection{Temporal pyramidal network}\label{temporal pyramidal}

Considering the similarity between consecutive frames and also the pattern of difficulty for this task, it leads to the idea of introducing adaptive joint processing. We applied this by proposing temporal pyramidal models. 

\begin{figure}[t!]
    \centering
    \begin{subfigure}{\linewidth}
    \includegraphics[width =\linewidth]{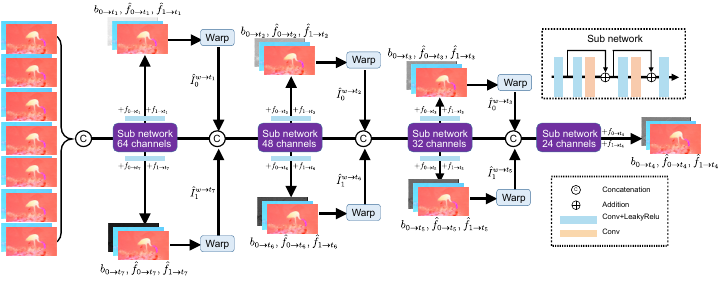}
    \caption{O.F. refinement network}
    \label{fig:of ref}
    \end{subfigure}
    \begin{subfigure}{\linewidth}
    \includegraphics[width =\linewidth]{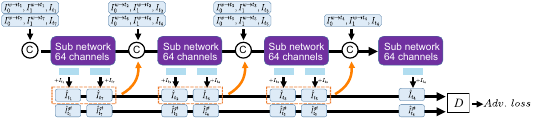}
    \caption{Post processing network}
    \label{fig:pixel ref}
    \end{subfigure}
    \caption{ The pyramidal network model designed for O.F. refinement (a) and adaptive pyramidal structure in post processing (b).
    }
    \label{fig: architeture}
\end{figure}

\noindent\textbf{Temporal pyramidal network for O.F. refinement}. The bidirectional O.F.s $f_{0 \rightarrow t_i}$ and $f_{1 \rightarrow t_i}$ predicted by~\eqref{eq:middle_of} are based on the O.F.s computed among the input frames. The initial prediction may inherit errors from flow estimation and cubic motion modeling, notably for the motion boundaries~\cite{jiang2018super}. To effectively improve $f_{0 \rightarrow t_i}$ and $f_{1 \rightarrow t_i}$, unlike the existing methods~\cite{DAIN,bao2018memc,liu2017video,jiang2018super,niklaus2018context,yuan2019zoom,peleg2019net,liu2019deep}, we aim to consider the relationship among intermediate frames and process all at one forward pass. To this end, we propose a temporal pyramidal O.F. refinement network, which enforces a strong bond between the intermediate frames, as shown in~\figurename~\ref{fig:of ref}. The network takes the concatenation of seven pairs of predicted O.F.s as input and adaptively refines the O.F.s based on the expected quality of the interpolation correspond to the distance to $I_0$ and $I_1$. In fact, the closest ones, $I_{t_1}$ and $I_{t_7}$ are processed only by one level of pyramid as they are more likely to achieve higher quality. With the same patterns, ($I_{t_2}$, $I_{t_6}$) are processed by two levels, ($I_{t_3}$, $I_{t_5}$) by three levels and finally $I_{t_4}$ by the entire four levels of the network as it is expected to achieve the lowest quality in interpolation.

To fully utilize the refined O.F.s, we warp $I_0$ and $I_1$ by the refined O.F. in each level as $I_0^{w\rightarrow t_i}$ and $I_1^{w\rightarrow t_i}$ and feed them to the next level. It is helpful to achieve better results in the next level as the warped frames are one step closer in time domain toward the locations in the target frame of that layer compared to $I_0$ and $I_1$. Thus, the motion between $I_0$ and $I_1$ is composed of step-wise motions, each measured within a short temporal interval. 

Additional to the refined O.F. at each level, a blending mask $b_{t_i}$~\cite{yuan2019zoom} is also generated. Therefore, the intermediate frames can be synthesized as~\cite{yuan2019zoom} by
\begin{equation}
    I_{t_i} = b_{t_i}\odot g(I_0, \hat{f}_{0 \rightarrow t_i}) + (1-b_{t_i}) \odot g(I_1, \hat{f}_{1 \rightarrow t_i}),
    \label{eq: frame synthesis}
\end{equation}
where $\hat{f}_{0 \rightarrow t_i}$ and $\hat{f}_{1 \rightarrow t_i}$ are refined bidirectional O.F. at $t_i$, $\odot$ denotes element-wise multiplication, and $g(\cdot, \cdot)$ is the bilinear warping function from~\cite{yuan2019zoom,jaderberg2015spatial}.

\noindent\textbf{Temporal pyramidal network for post processing}. The intermediate frames synthesized by~\eqref{eq: frame synthesis} may still contain artifacts due to the inaccurate O.F., blending masks, or synthesis process. Therefore, we introduce a post processing network following the similar idea of the O.F. refine network to adaptively refine the interpolated frames $I_{t_i}$. However, as the generated frames are not aligned, feeding all the frames at the beginning level cannot properly enhance the quality. Instead, we input the generated frame separately at different levels of the network according to the temporal distance, as shown in~\figurename~\ref{fig:pixel ref}. At each time stamp $t_i$, we also feed the warped inputs $I^{w \rightarrow t_i}_0$ and $I^{w \rightarrow t_i}_1$ to reduce the error caused by inaccurate blending masks. Similar to O.F. refinement network, the refined frames $\hat{I}_{t_i}$ are also fed to the next level as guidance.

For both pyramidal networks, we employ the same sub network for each level of the pyramid and adopt residual learning to learn the O.F. and frame residuals. The sub network is composed of two residual blocks proposed by~\cite{nah2017deep} and one convolutional layer at the input and another at the output. We set the number of channels in a reducing order for O.F. refinement pyramid, as fewer frames are dealt with when moving to the middle time step. In contrast, we keep the same channel numbers for all the levels of post processing module.

\subsection{Loss functions}
The proposed integrated network for multi-frame interpolation targets temporal consistency by joint optimization of all frames. To further impose consistency between frames, we apply generative adversarial learning scheme~\cite{zhang2019exploiting} and two-player min-max game idea in~\cite{goodfellow2014generative} to train a discriminator network $D$ which optimizes the following problem:
\begin{equation}
    \min_{G}\max_{D} \mathbb{E}_{\textbf{g}\sim p(I^{gt}_{t_i})}[\mbox{log}D (\textbf{g})] + \mathbb{E}_{\textbf{x}\sim p(I)}[\mbox{log}(1-D(G(\textbf{x})))],
\end{equation}
where $\textbf{g} = [I^{gt}_{t_1}, \cdots I^{gt}_{t_7}]$ are the seven ground truth frames and $\textbf{x}=[I_{-1}, I_0, I_1, I_2]$ are the four input frames. We add the following generative component of the GAN as the temporal loss~\cite{zhang2019exploiting,ledig2017photo}:
\begin{equation}
    \mathcal{L}_{temp} = \sum_{n=1}^{N}-\mbox{log}D(G(\textbf{x})).
\end{equation}
The proposed framework in~\figurename~\ref{fig:overall} serves as a generator and is trained alternatively with the discriminator.
To optimize the O.F. refinement and post processing networks, we apply the $\ell_1$ loss. The whole architecture is trained by combining all the loss functions:
\begin{equation}
    \mathcal{L} = \sum_{i=1}^{7}(\left \| \hat{I}_{t_i}-I^{gt}_{t_i} \right \|_1 + \left \| I_{t_i}-I^{gt}_{t_i} \right \|_1) + 
 \mathcal{L}_{w_{relax}} + \lambda\mathcal{L}_{temp},
 \label{eq:total loss}
\end{equation}
where the $\lambda$ is the weighting coefficient and equals to 0.001.

\section{Experiments}
In this section, we provide the implementation details and the analysis of the results of the proposed method in comparison to the other methods and different ablation studies.

\subsection{Implementation details}\label{implementation}
To train our network, we collected a dataset of 903 short video clips (2 to 10 seconds) with the frame rate of 240fps and a resolution of 720$\times$1280 from YouTube. The videos are covering various scenes, and we randomly select 50 videos for validation. From these videos, we created 8463 training samples of 25 consecutive frames as in~\cite{xu2019quadratic}. Our model takes the 1$^{st}$, 9$^{th}$, 17$^{th}$, and 25$^{th}$ frames as inputs to generate the seven frames between the 9$^{th}$ and 17$^{th}$ frames by considering 10$^{th}$ to 16$^{th}$ frames as ground truths. We randomly crop 352$\times$352 patches and apply horizontal, vertical as well as temporal flip for data augmentation in training.

To improve the convergence speed, a stage-wise training strategy is adopted \cite{zhang2018densely}. We first train each module except the discriminator using $\ell_1$ loss independently for 15 epochs with the learning rate of $10^{-4}$ by not updating other modules. The whole network is then jointly trained using~\eqref{eq:total loss} and learning rate of $10^{-5}$ for 100 epochs. We use the Adam optimizer~\cite{kingma2014adam} and empirically set the neighborhood range $d$ in~\eqref{eq:relaxation} to 9. During the training, the pixel values of all images are scaled to [-1, 1]. All the experiments are conducted on an Nvidia V100 GPU. More detailed network architecture will be provided in the supplementary material.


\subsection{Evaluation datasets}

We evaluate the performance of the proposed method on widely used datasets including two multi-frame interpolation dataset~(Adobe240~\cite{su2017deep} and GOPRO~\cite{nah2017deep}) and two single-frame interpolation~(Vimeo90K~\cite{xue2019video} and DAVIS\cite{perazzi2016benchmark}). Adobe240 and GOPRO are initially designed for deblurring tasks with a frame rate of 240fps and resolution of 720$\times$1280. Both are captured by hand-held high-speed cameras and contain a combination of object and camera motion in different levels, which makes them challenging for the frame interpolation task. We follow the same setting as Sec.~\ref{implementation} to extract 4276 and 1393 samples of frame patch for Adobe240 and GOPRO, respectively. DAVIS dataset is designed for video segmentation, which normally contains large motions. It has 90 videos, and we extract 2637 samples of 7 frames. As for Vimeo90K, since the interpolation sub-set only contains triplets, which are not applicable for our methods as we need more frames for cubic motion modeling. Instead, we use the super-resolution test set, which contains 7824 samples of 7 consecutive frames. We interpolate 7 frames for Adobe240 and GOPRO and interpolate the 4$^{th}$~(middle) frame for DAVIS and Vimeo90K by using the 1$^{st}$, 3$^{rd}$, 5$^{th}$ and 7$^{th}$ frames as inputs.
\begin{table*}[t!]
\centering
\caption{Performance evaluation of the proposed method compared to the state-of-the-art methods in different datasets.
}
\scriptsize
\setlength{\tabcolsep}{2.2pt} 
\begin{tabular}{lcccccccccccc}
\toprule
\multirow{2}{*}{Methods} & 
\multicolumn{3}{c}{Adobe240} & 
\multicolumn{3}{c}{GoPro} &
\multicolumn{3}{c}{Vimeo90K} & 
\multicolumn{3}{c}{DAVIS} \\
\cmidrule(r){2-4} \cmidrule(r){5-7} \cmidrule(r){8-10} \cmidrule(r){11-13} 
& PSNR & SSIM & TCC & PSNR & SSIM & TCC & PSNR & SSIM & IE & PSNR & SSIM & IE\\
\midrule
SepConv & 32.38 & 0.938 & 0.832 & 30.82 & 0.910 & 0.789 & 33.60 & 0.944 & 5.30 & 26.30 & 0.789 & 15.61 \\
Super SloMo & 31.63 & 0.927 & 0.809 & 30.50 & 0.904 & 0.784 & 33.38 & 0.938 & 5.41 & 26.00 & 0.770 & 16.19\\
DAIN & 31.36 & 0.932 & 0.808 & 29.74 & 0.900 & 0.759 & 34.54 & 0.950 & 4.76 & 27.25 & 0.820 & 13.17\\
Quadratic & 32.80 & 0.949 & 0.842 & 32.01 & 0.936 & 0.822 & 33.62 & 0.946 & 5.22 & 27.38 & 0.834 & 12.46 \\
Ours  & \textbf{34.37} & \textbf{0.959} & \textbf{0.860} & \textbf{32.91} & \textbf{0.943} & \textbf{0.837} & \textbf{34.93} & \textbf{0.951} & \textbf{4.70} & \textbf{27.91} & \textbf{0.837} & \textbf{12.40}\\

\bottomrule
\end{tabular}
\label{tab: compare SOTA}
\end{table*}

\begin{figure}[t]
    \centering
    \begin{tabular}{c|ccccc}
    \includegraphics[width =0.16\linewidth]{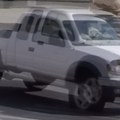}&
    \includegraphics[width =0.16\linewidth]{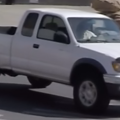}&
    \includegraphics[width =0.16\linewidth]{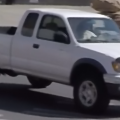}&
    \includegraphics[width =0.16\linewidth]{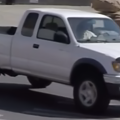}&
    \includegraphics[width =0.16\linewidth]{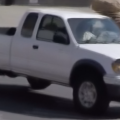}&
    \includegraphics[width =0.16\linewidth]{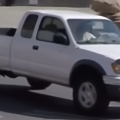}\\
    
    \includegraphics[width =0.16\linewidth]{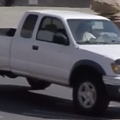}&
    \includegraphics[width =0.16\linewidth]{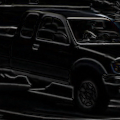}&
    \includegraphics[width =0.16\linewidth]{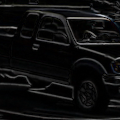}&
    \includegraphics[width =0.16\linewidth]{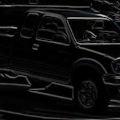}&
    \includegraphics[width =0.16\linewidth]{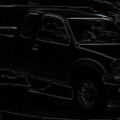}&
    \includegraphics[width =0.16\linewidth]{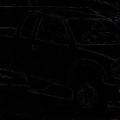}\\
    Input$\&$GT & SepConv & Super SloMo & DAIN & Qudratic & Ours
    \end{tabular}
    \caption{An example from Adobe240 to visualize the temporal consistency. The top row shows the middle frames generated by different methods, and the bottom row shows the interpolation error. Our method experiences less shifting in the temporal domain.
    }
    \label{fig: visualize temporal consistency}
\end{figure}

\subsection{Comparison with the state-of-the-arts}

We compare our method with four state-of-the-art frame interpolation methods: Super SloMo~\cite{jiang2018super}, Quadratic~\cite{xu2019quadratic}, DAIN~\cite{DAIN}, and SepConv~\cite{niklaus2017video}, where we train ~\cite{jiang2018super} and ~\cite{xu2019quadratic} on our training data and use the model released by authors in the last two. We use PSNR, SSIM and interpolation error~(IE)~\cite{baker2011database} as evaluation metrics. For multi-frame interpolation in GOPRO and Adobe240, we borrow the concept of Temporal Change Consistency~\cite{zhang2019exploiting} which compares the generated frames and ground truth in terms of changes between adjacent frames by

\begin{equation}
    TCC(F,G) = \frac{\sum_{i=1}^{6}\mbox{SSIM}(abs(f^i-f^{i+1}), abs(g^i-g^{i+1})) }{6},
\end{equation}
where, $F=(f^1,\cdots, f^7)$ and $G=(g^1,\cdots, g^7)$ are the 7 interpolated and ground truth frames respectively. For the multi-frame interpolation task, we report the average of the metrics for 7 interpolated frames. The results reported in Table~\ref{tab: compare SOTA}, shows that our proposed method consistently performs better than the existing methods on both single and multi-frame interpolation scenarios. Notably, for multi-frame interpolation datasets~(Adobe240 and GOPRO), our method significantly outperforms the best existing method~\cite{xu2019quadratic} by 1.57dB and 0.9dB. The proposed method also achieves the highest temporal consistency measured by TCC thanks to the temporal pyramid structure and joint optimization of the middle frames, which exploits the temporal relation among the middle frames.

\begin{table*}[t!]
\centering
\caption{Ablation studies on the network components on Adobe240 and GOPRO.
}
\scriptsize
\setlength{\tabcolsep}{2.4pt} 
\begin{tabular}{lcccccccccc}
\toprule
\multirow{2}{*}{Methods} & 
\multicolumn{4}{c}{Adobe240} & 
\multicolumn{4}{c}{GOPRO} \\
\cmidrule(r){2-5} \cmidrule(r){6-9} 
& PSNR & SSIM & IE & TCC & PSNR & SSIM & IE & TCC\\
\midrule
w/o post pro.. & 33.87 & 0.954 & 6.21 & 0.848 & 32.63 & 0.942 & 6.80 & 0.831\\
w/o adv. loss & 34.35 & 0.958 & 5.89 & 0.850 & 32.86 & 0.942 & 6.77 & 0.830\\
w/o 2$^{nd}$ O.F. & 34.24 & 0.957 & 5.97 & 0.854 & 32.73 & 0.940 & 6.91 & 0.832\\
w/o O.F. relax. & 33.92 & 0.955 & 6.14 & 0.851 & 32.45 & 0.936 & 7.09 & 0.828\\
w/o pyr. & 33.92 & 0.954 & 6.33 & 0.845 & 32.37 & 0.935 & 7.30 & 0.820\\
Full model  & \textbf{34.37} & \textbf{0.959} & \textbf{5.89} & \textbf{0.860} & \textbf{32.91} & \textbf{0.943} & \textbf{6.74} & \textbf{0.837}\\
\bottomrule
\end{tabular}
\label{tab: ablation component}
\end{table*}
\begin{figure}[t]
\centering
\begin{tabular}{cccccccc}
	\centering 
	\rotatebox{90}{\cite{jiang2018super}} & 
	\includegraphics[width =0.12\linewidth]{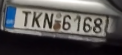} &
	\includegraphics[width =0.12\linewidth]{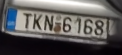} &
	\includegraphics[width =0.12\linewidth]{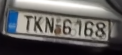} &
	\includegraphics[width =0.12\linewidth]{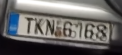} &
	\includegraphics[width =0.12\linewidth]{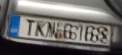} &
	\includegraphics[width =0.12\linewidth]{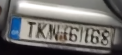} &
	\includegraphics[width =0.12\linewidth]{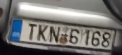} \\
	
	\rotatebox{90}{\cite{xu2019quadratic}} & 
	\includegraphics[width =0.12\linewidth]{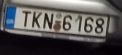} &
	\includegraphics[width =0.12\linewidth]{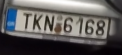} &
	\includegraphics[width =0.12\linewidth]{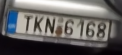} &
	\includegraphics[width =0.12\linewidth]{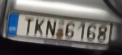} &
	\includegraphics[width =0.12\linewidth]{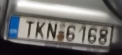} &
	\includegraphics[width =0.12\linewidth]{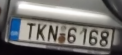} &
	\includegraphics[width =0.12\linewidth]{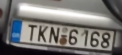} \\
	
	\rotatebox{90}{Ours} & 
	\includegraphics[width =0.12\linewidth]{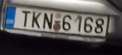} &
	\includegraphics[width =0.12\linewidth]{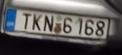} &
	\includegraphics[width =0.12\linewidth]{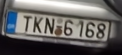} &
	\includegraphics[width =0.12\linewidth]{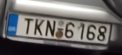} &
	\includegraphics[width =0.12\linewidth]{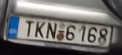} &
	\includegraphics[width =0.12\linewidth]{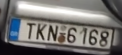} &
	\includegraphics[width =0.12\linewidth]{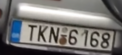} \\
	
	\rotatebox{90}{GT}& 
	\includegraphics[width =0.12\linewidth]{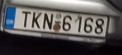} &
	\includegraphics[width =0.12\linewidth]{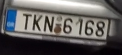} &
	\includegraphics[width =0.12\linewidth]{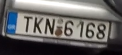} &
	\includegraphics[width =0.12\linewidth]{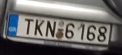} &
	\includegraphics[width =0.12\linewidth]{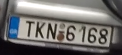} &
	\includegraphics[width =0.12\linewidth]{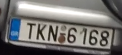} &
	\includegraphics[width =0.12\linewidth]{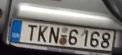} \\
	& $I_{t_1}$ & $I_{t_2}$ & $I_{t_3}$ & $I_{t_4}$ & $I_{t_5}$ & $I_{t_6}$ & $I_{t_7}$ 
	
\end{tabular}
\caption{
Visualization of the seven intermediate frames of $I_{t_{1}}$ to $I_{t_{7}}$ generated by our method compared to Quadratic~\cite{xu2019quadratic} and Super~SloMo~\cite{jiang2018super} from GOPRO.
}
\label{fig: temporal adaptiveness}
\end{figure}

In addition to the TCC, to better show the power of the proposed method in preserving temporal consistency between frames, \figurename~\ref{fig: visualize temporal consistency} reports $\hat{I}_{t_4}$ and IE generated by different methods from Adobe240. As shown in~\figurename~\ref{fig: visualize temporal consistency}, the generated middle frames by different methods are visually very similar to the ground truth. However, a comparison of the IE reveals significant errors that occurred near the edges of moving objects caused because of time inconsistency between generated frames and the ground truth. In contrast, our method generates a high-quality consistent frame with the ground truth in both spatial and temporal domains.

Another example from GOPRO in~\figurename~\ref{fig: temporal adaptiveness}, shows the results of the proposed method in comparison with Super SloMo~\cite{jiang2018super} and Quadratic~\cite{xu2019quadratic} which they have not applied any adaptive processing for frames interpolation. As it can be seen in~\figurename~\ref{fig: temporal adaptiveness}, at $t_1$ and $t_7$ which are closer to the input frames, all the methods generate comparable results. However, approaching to the middle frame as the temporal distance from the input increases, the quality of frames generated by Super SloMo and Quadratic start to degrade while our method experiences less degradation and higher quality. Especially for $I_{t_4}$, our improvement is significant, as also shown by the PSNR values at each time stamp $t_i$ in~\figurename~\ref{fig: 7 psnr}.

\begin{figure}[t!]
    \centering
    \begin{subfigure}{0.20\linewidth}
    \includegraphics[width =\linewidth]{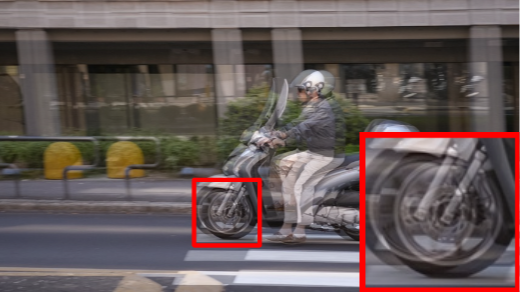}
    \caption*{Inputs\\ \quad}
    \end{subfigure}
    \begin{subfigure}{0.11\linewidth}
    \includegraphics[width =\linewidth]{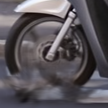}
    \caption*{SepConv\\ \quad}
    \end{subfigure}
    \begin{subfigure}{0.11\linewidth}
    \includegraphics[width =\linewidth]{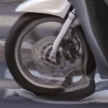}
    \caption*{\centering Super SloMo}
    \end{subfigure}
    \begin{subfigure}{0.11\linewidth}
    \includegraphics[width =\linewidth]{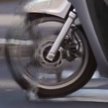}
    \caption*{DAIN\\ \quad}
    \end{subfigure}
    \begin{subfigure}{0.11\linewidth}
    \includegraphics[width =\linewidth]{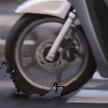}
    \caption*{Quadratic\\ \quad}
    \end{subfigure}
    \begin{subfigure}{0.11\linewidth}
    \includegraphics[width =\linewidth]{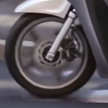}
    \caption*{Ours\\ \quad}
    \end{subfigure}
    \begin{subfigure}{0.11\linewidth}
    \includegraphics[width =\linewidth]{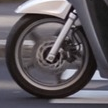}
    \caption*{GT\\ \quad}
    \end{subfigure}
    \caption{Sample results for interpolating the middle frame for a complex motion example from DAVIS dataset.} 
    \label{fig: compare DAVIS}
\end{figure}

Our method also works better on DAVIS and Vimeo90K, as reported in Table~\ref{tab: compare SOTA}. \figurename~\ref{fig: compare DAVIS} shows an example of a challenging scenario that involves both translational and rotational motion. The acceleration-aware Quadratic can better estimate the motion, while others have undergone severe degradation. However, undesired artifacts are still generated by Quadratic near the motion boundary. In contrast, our method exploits the cubic motion modeling and temporal pyramidal processing, which better captures this complex motion and generates comparable results against the ground truth.

\subsection{Ablation studies}\label{ablation section}

\textbf{Analysis of the model.} 
To explore the impact of different components of the proposed model, we investigate the performance of our solution when applying different variations including 1) w/o post pro.: removing post processing; 2) w/o adv. loss: removing adversarial loss; 3) w/o 2$^{nd}$ O.F.: replace the second stage flow estimation with the exact same network as the first stage; 4) w/o O.F. relax.: replace $\mathcal{L}_{w_{relax}}$ by $\mathcal{L}_{\ell_1}$; 5) w/o pyr.: in both pyramidal modules, we place all the input as the first level of the network, and the outputs are caught at the last layer. The performance of the above variations evaluated on Adobe240 and GOPRO datasets, as shown in Table~\ref{tab: ablation component}, reveals that all the listed modifications lead to degradation in performance. As expected, motion relaxation and the pyramidal structure are important as they provide more accurate motion prediction and enforce the temporal consistency among the interpolated frames, as reflected in TCC. The post processing as its missing in the model also brings a large degradation is a crucial component that compensates the inaccurate O.F. and blending process. It is worth noting that even though the quantitative improvement of PSNR and SSIM for the adversarial loss is small, it is effective to preserve the temporal consistency as reported by the TCC values.

\begin{table*}[t!]
\centering
\scriptsize
\setlength{\tabcolsep}{2.4pt} 
\begin{minipage}{.47\linewidth}
\centering
\caption{\small Comparison between linear, quadratic and cubic motion models. }
\label{tab: motion order comparison}
\begin{tabular}{lcccccc}
\toprule
\multirow{2}{*}{Models} & 
\multicolumn{3}{c}{Adobe240} & 
\multicolumn{3}{c}{GOPRO} \\
\cmidrule(r){2-4} \cmidrule(r){5-7} 
& PSNR & SSIM & IE & PSNR & SSIM & IE \\
\midrule
Linear & 33.97 & 0.955 & 6.13 & 32.40 & 0.936 & 7.09 \\
Quad. & 34.24 & 0.957 & 5.95 & 32.70 & 0.941 & 6.85\\
Cubic  & \textbf{34.37} & \textbf{0.959} & \textbf{5.89} & \textbf{32.91} & \textbf{0.943} & \textbf{6.74} \\
\bottomrule
\end{tabular}
\end{minipage}\hfill
\begin{minipage}{.47\linewidth}
\centering
\caption{\small Comparison between models generating different number of frames.}
\label{tab: different number of frames}
\begin{tabular}{lcccc}
\toprule
\multirow{2}{*}{Methods} & 
\multicolumn{2}{c}{DAVIS} & 
\multicolumn{2}{c}{Vimeo90K} \\
\cmidrule(r){2-3} \cmidrule(r){4-5} 
& PSNR & SSIM & PSNR & SSIM \\
\midrule
1 frames & 27.07 & 0.819 & 32.02 & 0.944 \\
3 frames & 27.44 & 0.816 & 34.67 & 0.950\\
7 frames(no pyr.) & 27.25 & 0.815 & 34.56 & 0.950\\
7 frames & \textbf{27.91} & \textbf{0.837} & \textbf{34.93} & \textbf{0.951}\\
\bottomrule
\end{tabular}
\end{minipage}
\label{tab: ablation motion model}
\end{table*}
\textbf{Motion models.} To investigate the impact of different motion models, we trained our method with linear and quadratic~\cite{xu2019quadratic} motion prediction as well. The reported average quality in Table~\ref{tab: motion order comparison}, shows that the cubic modeling has been dominant in both GOPRO and Adobe240. Importantly, the improvement by quadratic against linear in the model proposed in~\cite{xu2019quadratic}, is reported to be more than 1dB, however, we observed 0.27dB and 0.3dB on Adobe240 and GOPRO datasets. We give credit to the proposed temporal pyramidal processing and applying motion relaxation. In comparison with the impact of quadratic over linear, our cubic modeling adds another 0.13dB and 0.21dB improvement on the Adobe240 and GOPRO, respectively, which shows the necessity of applying cubic modeling on the complexity of motions we have in different videos.

\begin{table*}[t!]
\centering
\caption{Motion relaxation evaluation for warping, prediction and final results.}
\scriptsize
\setlength{\tabcolsep}{2.4pt} 
\begin{tabular}{lccccccc}
\toprule
\multirow{2}{*}{Datasets} & 
\multicolumn{2}{c}{PSNR($I^{w\rightarrow 0}_1$, $I_0$)} & 
\multicolumn{2}{c}{PSNR($I^{w\rightarrow t_4}_1$, $I^{gt}_{t_4}$)} &
\multicolumn{2}{c}{PSNR($\hat{I}_{t_4}$, $I^{gt}_{t_4}$))}\\
\cmidrule(r){2-3} \cmidrule(r){4-5} \cmidrule(r){6-7} 
& $\mathcal{L}_{\ell_1}$ & $\mathcal{L}_{w_{relax}}$ & $\mathcal{L}_{\ell_1}$ & $\mathcal{L}_{w_{relax}}$ & $\mathcal{L}_{\ell_1}$ & $\mathcal{L}_{w_{relax}}$   \\
\midrule
DAVIS & \textbf{30.13} & 23.37 & 25.13 & \textbf{25.43} & 27.15 & \textbf{27.91}\\
\bottomrule
\end{tabular}
\label{tab: evaluation of motion relaxation.}
\end{table*}

\begin{figure}[t]
    
    \centering
    \begin{subfigure}{0.19\linewidth}
    \includegraphics[width =\linewidth]{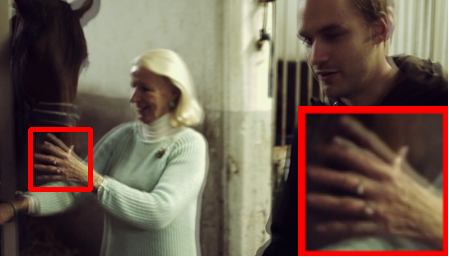}
    \end{subfigure}
    \begin{subfigure}{0.19\linewidth}
    \includegraphics[width =\linewidth]{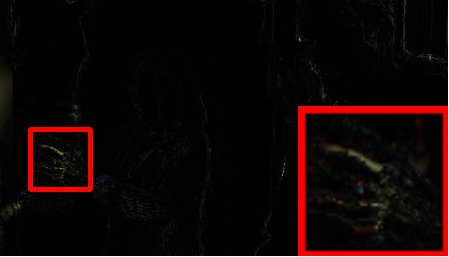}
    \end{subfigure}
    \begin{subfigure}{0.19\linewidth}
    \includegraphics[width =\linewidth]{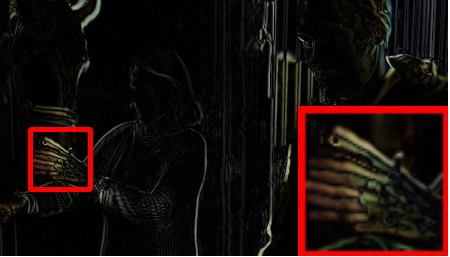}
    \end{subfigure}
    \begin{subfigure}{0.19\linewidth}
    \includegraphics[width =\linewidth]{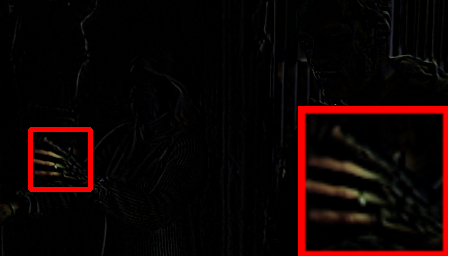}
    \end{subfigure}
    \begin{subfigure}{0.19\linewidth}
    \includegraphics[width =\linewidth]{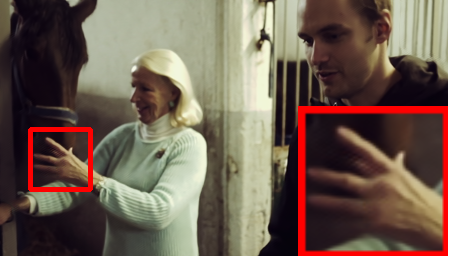}
    \end{subfigure}
    
    \centering
    \begin{subfigure}{0.19\linewidth}
    \includegraphics[width =\linewidth]{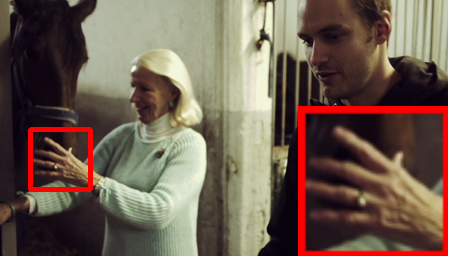}
    \caption*{Inputs $\&$ GT}
    \end{subfigure}
    \begin{subfigure}{0.19\linewidth}
    \includegraphics[width =\linewidth]{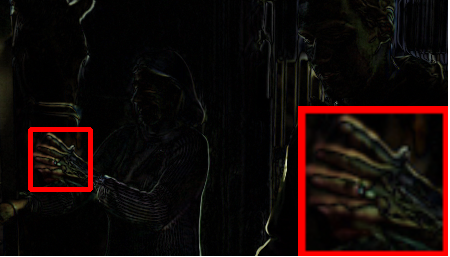}
    \caption*{Error~($I^{w \rightarrow 0}_1$)}
    \end{subfigure}
    \begin{subfigure}{0.19\linewidth}
    \includegraphics[width =\linewidth]{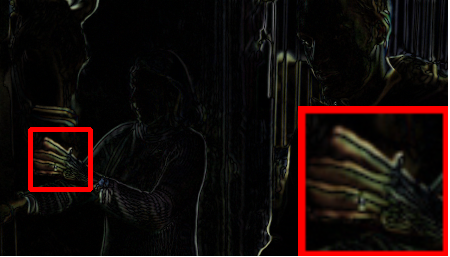}
    \caption*{Error~($I^{w \rightarrow t_4}_1$)}
    \end{subfigure}
    \begin{subfigure}{0.19\linewidth}
    \includegraphics[width =\linewidth]{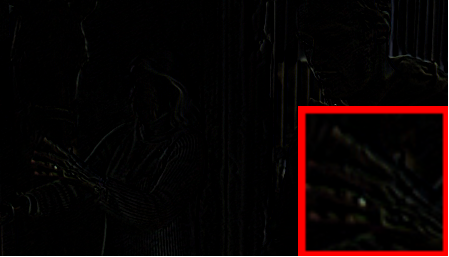}
    \caption*{Error~($\hat{I}_{t_4}$)}
    \end{subfigure}
    \begin{subfigure}{0.19\linewidth}
    \includegraphics[width =\linewidth]{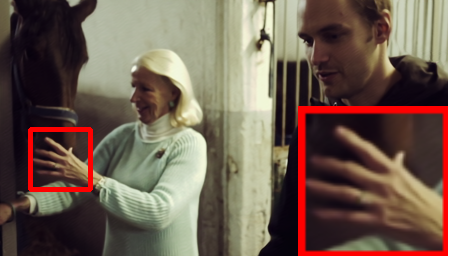}
    \caption*{$\hat{I}_{t_4}$}
    \end{subfigure}
    \caption{Sample results from Vimeo90K to show the comparison between O.F. estimation with (bottom row) and without (top row) relaxation in terms of the interpolation error for motion prediction and final interpolation result.
    }
    \label{fig: motion relaxation error}
\end{figure}

\textbf{Constraints relaxation in motion estimation.} To investigate the impact of applying motion estimation relaxation in our architecture, we train two versions of the entire solution, with relaxation ($\ell_{w_{relax}}$) and without relaxation ($\ell_1 $). For each case we perform three comparisons, first, $I_1$ warped by $f_{1 \rightarrow 0}$ which named ~($I^{w\rightarrow 0}_1$) and compare to $I_0$, second, $I_1$ warped by the predicted $f_{1 \rightarrow {t_4}}$ (before refinement) named by ($I^{w\rightarrow {t_4}}_1$) and compared to $I^{gt}_{t_4}$, and finally, we also compared the final output of the network with $I^{gt}_{t_4}$. Table~\ref{tab: evaluation of motion relaxation.} reports results of evaluation on DAVIS and \figurename~\ref{fig: motion relaxation error} shows IE for an example from Vimeo90k. 
Both Table~\ref{tab: evaluation of motion relaxation.} and~\figurename~\ref{fig: motion relaxation error} show that although the relaxation makes the O.F. estimation between two initial pair poor, it gives better initial motion prediction for the middle frame as well as the final interpolation result.

\textbf{Temporal pyramidal structure.}
The effectiveness of the temporal pyramidal structure in interpolating multiple frames has already been verified in Table~\ref{tab: ablation component}. To further investigate this impact by also considering the number of frames it generates, we trained another 3 variations of model including predicting all 7 frames without pyramidal structure, predicting 3 frames~($i=2,4,6$), and only 1 middle frame~($i=4$) with pyramidal model. Table~\ref{tab: different number of frames} reports the interpolation quality of the middle frame on DAVIS and Vimeo90K for all these cases. The results in Table~\ref{tab: different number of frames} demonstrate that the interpolation of the middle frame benefits from the joint optimization with other frames. 

\subsection{Efficiency analysis} 
\begin{figure}[t]
    \centering
    \begin{subfigure}{0.32\linewidth}
    \includegraphics[width =\linewidth]{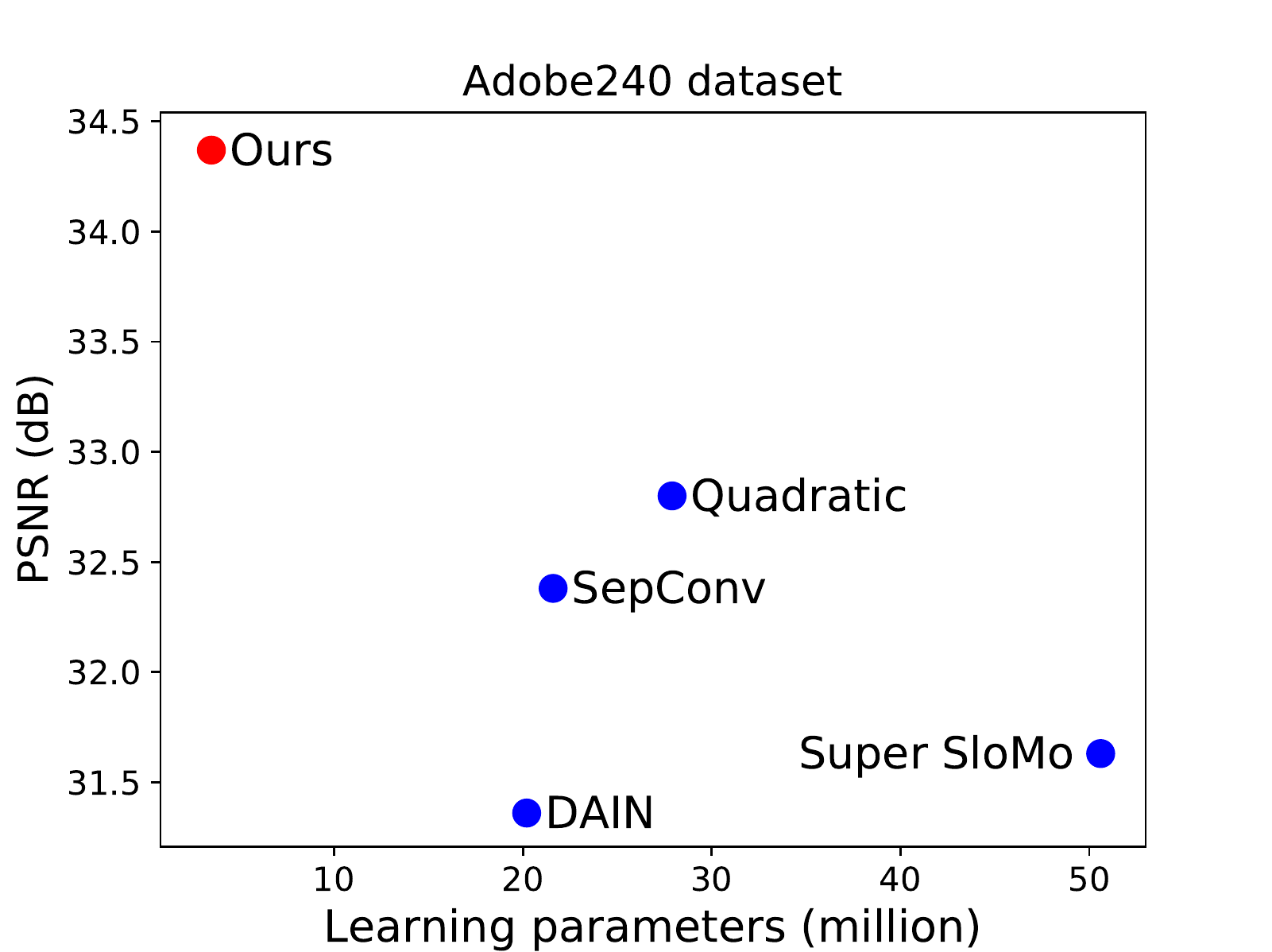}
    \caption{Model size VS. PSNR.}
    \label{fig: model size}
    \end{subfigure}
    \begin{subfigure}{0.32\linewidth}
    \includegraphics[width =\linewidth]{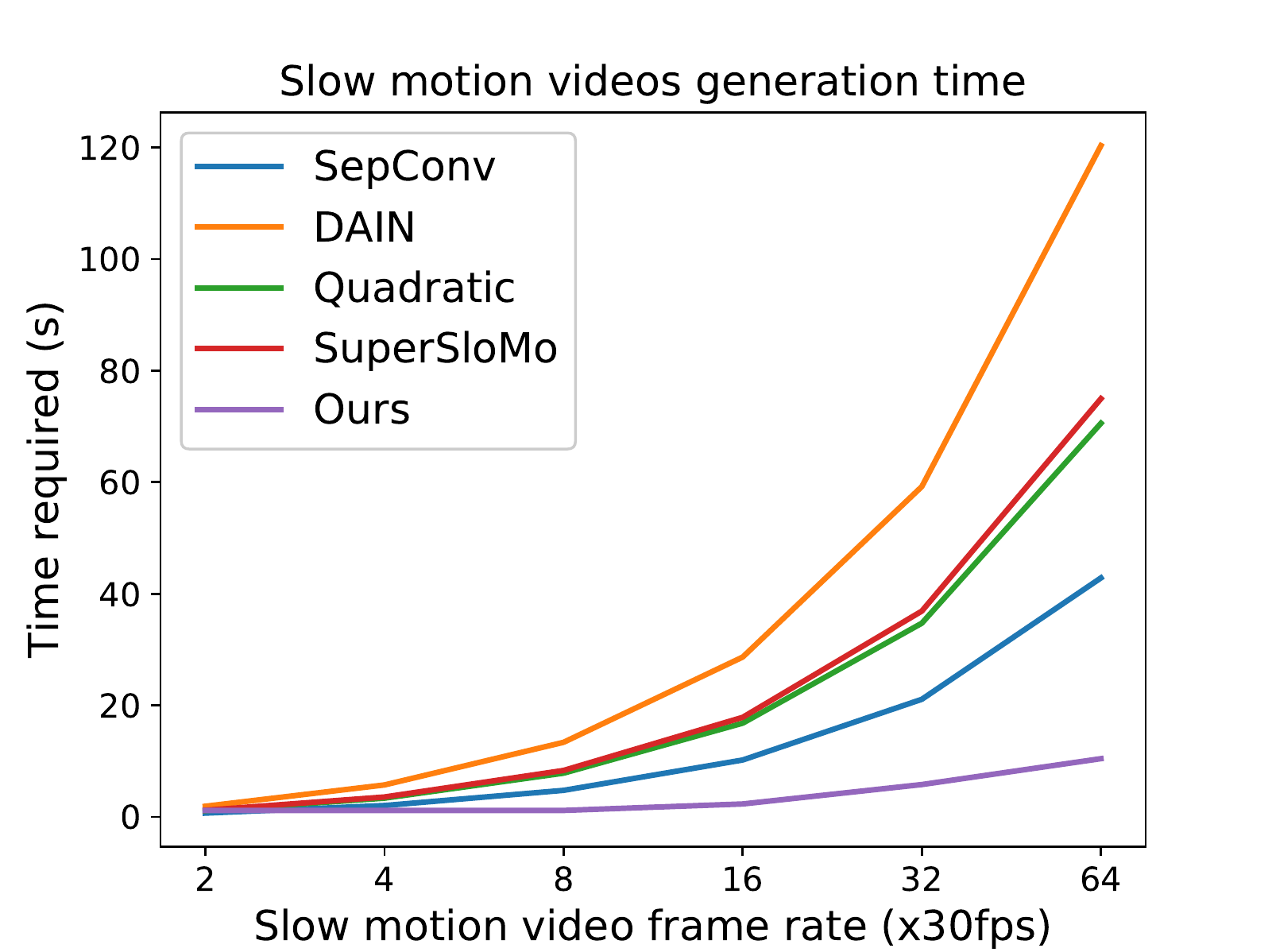}
    \caption{Inference speed.}
    \label{fig: speed}
    \end{subfigure}
    \begin{subfigure}{0.32\linewidth} 
    \includegraphics[width=\linewidth]{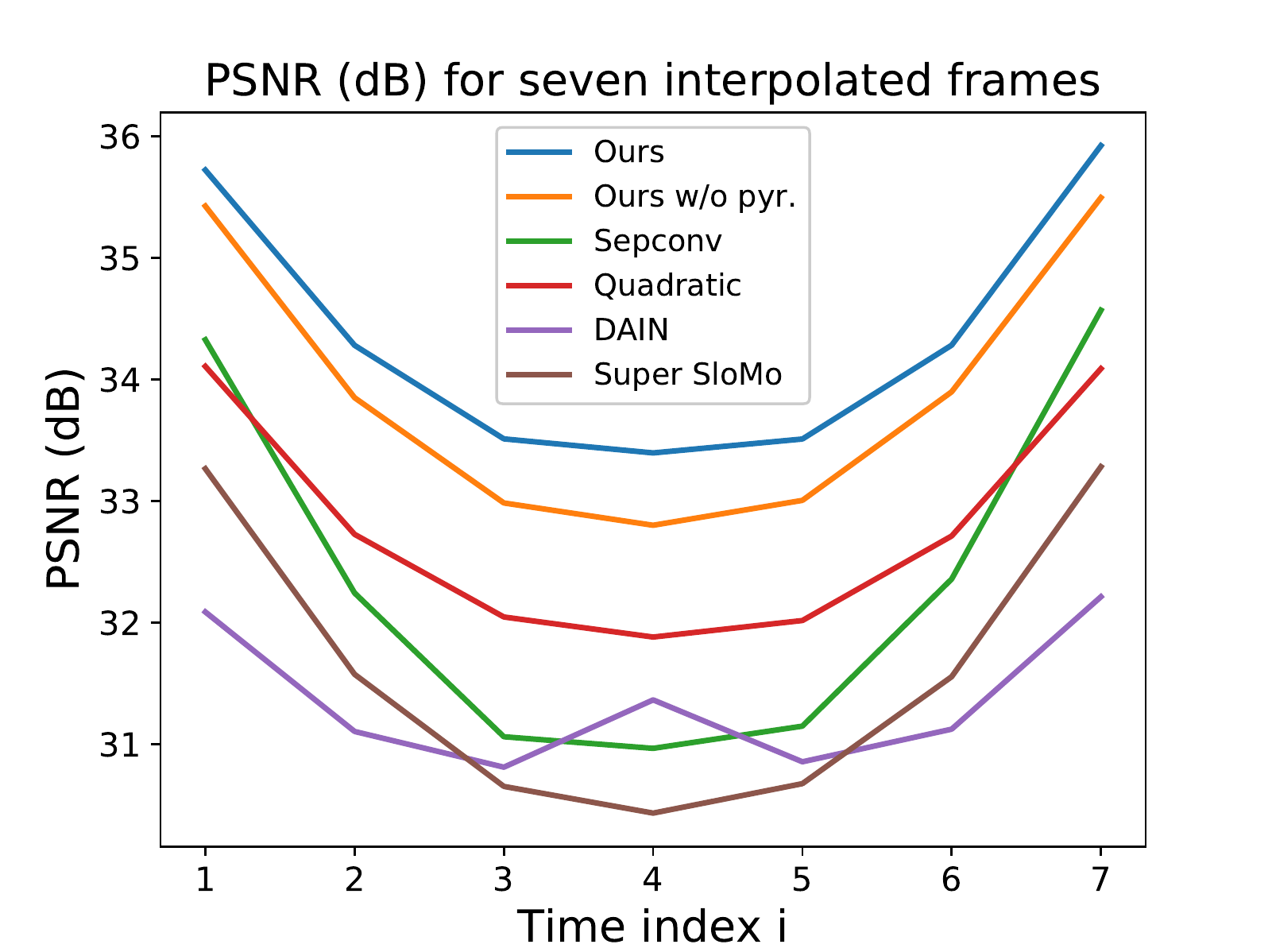}
    \caption{PSNR for 7 frames.}
    \label{fig: 7 psnr}
    \end{subfigure}
    \caption{Efficiency of the proposed method compared to state-of-the-art methods from the perspective of performance and model size (a), inference speed (b), and performance trend in multiple frame interpolation (c). 
    }
    \label{fig:efficiency}
\end{figure}

Considering the wide applications for frame interpolation, especially on mobile and embedded devices, investigating the efficiency of the solution is crucial. We report the efficiency of the proposed method in terms of model size, interpolation quality, and inference time. \figurename~\ref{fig: model size} reports PSNR values evaluated on Adobe240 in relation with the model sizes. The proposed method outperforms all the methods in the quality of the results with a large margin while having a significantly smaller model size. In particular, our method outperforms Quadratic~\cite{xu2019quadratic} by 1.57dB by using only 12.5$\%$ of its parameters. We also show the inference times for interpolating different numbers of frames in~\figurename~\ref{fig: speed}. To interpolate more than 8 frames, our method is able to be extended to interpolate more frames by simply adding more levels in the pyramid. However, higher frame rate videos are hard to be obtained for training; thus, we adopt the iterative interpolation method (run 8x model multiple times and drop the corresponding frames). As reported in~\figurename~\ref{fig: speed}, our method is around 7 times faster than~\cite{xu2019quadratic} for interpolating more than 8 frames. Our method is the fastest and has the smallest size while keeping the high-quality results for multi-frame interpolation tasks, which makes it applicable for low power devices. 

\section{Conclusions}

In this work, we proposed a powerful and efficient multi-frame interpolation solution that considers prior information and the challenges in this particular task. The prior information about the difficulty levels among the intermediate frames helps us to design a temporal pyramidal processing structure. To handle the challenges of real world complex motion, our method benefits from the proposed advanced motion modeling, including cubic motion prediction and relaxed loss function for flow estimation. All these parts together help to integrate multi-frame generation in a single optimized and efficient network while the temporal consistency of frames and spatial quality are at maximum level beating the state-of-the-art solutions.


\clearpage
%
%
\bibliographystyle{splncs04}
\bibliography{egbib}
\end{document}